# CLASSIFICATION OF BUILDINGS' POTENTIAL FOR SEISMIC DAMAGE BY MEANS OF ARTIFICIAL INTELLIGENCE TECHNIQUES

Konstantinos Kostinakis[1], Konstantinos Morfidis[2], Konstantinos Demertzis[3,4], Lazaros Iliadis[4]

[1]Assistant Professor, Department of Civil Engineering, Aristotle University of Thessaloniki
Aristotle University campus, 54124, Thessaloniki, Greece
e-mail: kkostina@civil.auth.gr

[2]Assistant Researcher, Earthquake Planning and Protection Organization (EPPO-ITSAK)
Terma Dasylliou, 55535, Thessaloniki, Greece
e-mail: konmorf@gmail.com

[3]Laboratory of Complex Systems, Department of Physics, Faculty of Sciences, International Hellenic University,
Kavala Campus, St. Loukas, 65404, Greece
e-mail: kdemertzis@teiemt.gr

[4]School of Engineering, Department of Civil Engineering, Faculty of Mathematics Programming and General Courses, Democritus University of Thrace, Kimmeria, Xanthi, Greece
e-mail: liliadis@civil.duth.gr

**Abstract.** Developing a rapid, but also reliable and efficient, method for classifying the seismic damage potential of buildings constructed in countries with regions of high seismicity is always at the forefront of modern scientific research. Such a technique would be essential for estimating the pre-seismic vulnerability of the buildings, so that the authorities will be able to develop earthquake safety plans for seismic rehabilitation of the highly earthquake-susceptible structures. In the last decades, several researchers have proposed such procedures, some of which were adopted by seismic code guidelines. These procedures usually utilize methods based either on simple calculations or on the application of statistics theory. Recently, the increase of the computers' power has led to the development of modern statistical methods based on the adoption of Machine Learning algorithms. These methods have been shown to be useful for predicting seismic performance and classifying structural damage level by means of extracting patterns from data collected via various sources. The present paper attempts to compare and evaluate the capability of various Machine Learning methods to adequately classify the seismic damage potential of R/C buildings. A large training dataset is used for the implementation of the classification algorithms. To this end, 90 3D R/C buildings with three different masonry infills' distributions are analysed utilizing Nonlinear Time History Analysis method for 65 real seismic records. The level of the seismic damage is expressed in terms of the Maximum Interstory Drift Ratio. A large number of Machine Learning algorithms is utilized in order to estimate the buildings' damage response. The most significant conclusion which is extracted is that the Machine Learning methods that are mathematically well-established and their operations that are clearly interpretable step by step can be used to solve some of the most sophisticated real-world problems in consideration with high accuracy.

## 1. Introduction

A large number of existing buildings were constructed in countries with regions of high seismicity. These structures were designed using older seismic codes that did not incorporate the most recent earthquake-resistant provisions, thus leading to high seismic vulnerability under earthquake excitations. For these buildings it is especially crucial to develop a rapid, but also reliable and efficient, method for classifying the seismic damage potential and for prioritizing the buildings with high seismic vulnerability, so that the authorities will be able to develop appropriate earthquake safety plans for seismic rehabilitation. Since now, several researchers have proposed such procedures, some of which were adopted by seismic code guidelines (e.g., see [1-8]). The most of these codes utilize simplified procedures in order to assess the seismic response and the structural damage level, based on certain input parameters such as structural configuration and seismic motion intensity measures. Additionally, to these methods, a number of researchers have developed techniques for the rapid estimation of the buildings' seismic vulnerability based on the application of statistical theory, e.g., seismic fragility curves (e.g. [9-15]). These techniques are characterized by certain shortcomings (small number of input structural and seismic parameters, linear relationship between inputs and outputs, simple formulae for the estimation of the damage level based on the input variables), which make their use rather limited and, in many cases, not effective, as they are not able to capture the full complexity of the relationship between damage and input parameters. In order to overcome the abovementioned limitations, in the last decades, modern statistical methods based on the adoption of Machine Learning (ML) algorithms were developed (e.g. [16-19]). The up-to-date research on these methods has shown that they can provide a fast, reliable, and computationally easy way for classify the buildings' the seismic damage potential and that they can successfully identify structural performance under seismic motions by extracting patterns from data collected via various sources. Machine learning is one of the most important and widespread fields of artificial intelligence that includes those computational methods of studying and constructing algorithms which can learn from appropriate datasets. Their success is based on the thorough processing of the data that record the behavior of a system, so that by detecting the appropriate patterns valuable information can be extracted. Based on this experience the ML algorithms are able to make accurate future predictions. The concept of experience refers to the hidden knowledge contained in the data collected from the field and related to the type of damage suffered by the buildings under investigation. In recent years it has been proven that ML algorithms have the ability to be successfully applied in many areas of modeling engineering problems, giving a serious breakthrough to modern earthquake engineering.

More specifically, several research studies have found that ML methods, mainly Artificial Neural Networks (ANNs), can effectively assess the seismic response of complex structures. A thorough literature review of the most commonly used and recently proposed ML methods for the buildings' seismic damage assessment has been made by Harirchian et. al [20], by Xie et al [21] and Sun et al [22]. Next, a brief review of some of the most significant relevant researches is given. Rafiq et al [23] adopted several different types of ANNs (Multi-layer Perceptron, Radial Basis Networks and normalized Radial basis Networks) in order to solve engineering problems. Aoki et al. [24] tried to assess the seismic vulnerability of chemical industrial plants with different topologies with the aid of probabilistic ANNs. In another research study, Lautour and Omenzetter [25] investigated 2D reinforced concrete frames that varied in topology, stiffness, strength and damping and were subjected to a suite of ground motions. They established the ability of the ANNs to reliably estimate the earthquake-induced damage level of these structures. Tesfamariam and Liu [26] studied eight different statistical damage classification techniques in order to estimate the reported seismic induced damage and proved the feasibility and effectiveness of the selected statistical approaches to classify the damage of R/C buildings. Similarly, Arslan [27] created a dataset for the training of ANNs by means of incremental static pushover analyses in order to estimate the ANNs' ability to predict the seismic damage level of medium and high-rise R/C buildings. Kia and Sensoy [28] used a combination of ANNs with SVM in order to classify the damage of R/C slab-column frames. Kostinakis and Morfidis carried out a number of research studies [30-34] in order to assess the efficiency of ANNs regarding the classification of R/C buildings' seismic damage. The same authors investigated also the number and the combination of the input parameters through which an optimum prediction for the damage state of R/C buildings can be achieved, the influence of the parameters which are used for the configuration of the networks' training on the efficiency of their predictions, as well as the impact of the presence of masonry infills on the results. More recently, Zhang et al. [35] used predictive models including classification and regression tree and Random Forests in order to probabilistically identify the structural safety state of an earthquake-damaged building. The same research team, in another research work [36], adopted several ML techniques for the adequate estimation of the residual structural capacity of damaged tall buildings. A different approach was given by Harirchian E. and Lahmer T [37], who developed a novel method based on type-2 fuzzy for earthquake vulnerability assessment of buildings via Rapid Visual Screening. Mangalathu et. al. [38], using data from the 2014 South Napa earthquake, examined the ability of ML methods, such as discriminant analysis, k-nearest neighbors, decision trees, and random forests, to rapidly estimate seismic building damage. The same research team conducted also a number of works [39-42] in an attempt to thoroughly investigate the applicability of a series of ML techniques to predict the potential of structures for earthquake-induced damage. Similar scientific investigation was conducted by Harirchian and Lahmer and their research team [43-47]. The results of the most research works established the capability of ML methods in the successful seismic damage classification of structures. However, there is a rather limited number of researches that used a large number of ML methods, structures and seismic motions in order to comparatively evaluate the ML techniques' efficiency in estimating the seismic damage response with adequate reliability.

In an attempt to further investigate the feasibility of adopting Machine Learning methods for the estimation of the earthquake-induced damage potential, the present paper attempts a comparative evaluation of a large number of Machine Learning algorithms for the reliable classification of R/C buildings' potential for seismic damage. For this aim, a training dataset consisting of 30 3D R/C buildings with different structural parameters was chosen. The buildings were designed based on the provisions of EC2 [48] and EC8 [49]. For each one of these buildings three different configurations as regards their masonry infills were considered (without masonry infills, with masonry infills in all stories and with masonry infills in all stories except for the ground story), leading to three different data subsets with 30 buildings each. Then, the buildings were analysed my means of the Nonlinear Time History Analyses method (NTHA) for 65 appropriately chosen real earthquake records. Both seismic and structural parameters widely used in the literature were selected as inputs in the process of Machine Learning methods. The quantification of the buildings' damage level was done by means of the well-documented Maximum Interstory Drift Ratio (MIDR). The methodology of the proposed assessment/information system uses and extends the most technologically advanced methods of forecasting, analysis and modeling of seismic engineering, as it extracts the hidden knowledge found in digital data, in order to add intelligence to the best decision support methods. At the same time, it gives the stimulus for the utilization of intelligent methods and their penetration in the development sector, for huge innovative leaps and development of activities that were previously impossible.

## 2. Dataset generation

A large training dataset consisting of buildings with different structural characteristics was used to generate the database for the training and testing of the ML models. The structures have characteristics that are common to buildings designed and built on the basis of modern seismic codes and according to the construction practice in most european countries with regions of high seismicity. In particular, 30 R/C buildings with structural systems consisting of members in two perpendicular directions (axes x and y) were selected. The buildings are rectangular in plan and regular in elevation and in plan according to the criteria set by EC8 [49]. The structures differ in the total height $H_{tot}$ ($H_{tot}$ = (stories' number) x (stories' height: 3.2m)), the value of structural eccentricity $e_{cc\_tot}$ (i.e., the distance between the mass center and the stiffness center of stories) and the ratio of the base shear received by the walls along two horizontal orthogonal directions (axes x and y): $V_{w1}$ and $V_{w2}$. A detailed description of the investigated buildings can be found in [31]. The influence of the masonry infill walls, the placement of which along the height of the buildings is part of the traditional building practice, on the structures' seismic response and damage was considered taking into account for each one of the 30 structures three different assumptions about their distribution. More specifically, three different training subsets were generated: (a) subset denoted as ROW_FORM_BARE consisting of the 30 buildings without masonry infills (bare structures), (b) subset denoted as ROW_FORM_FULL-MASONRY consisting of the 30 buildings with masonry infills uniformly distributed along the height (infilled structures) and (c) subset denoted as ROW_FORM_PILOTIS consisting of the 30 buildings with the first story bare and the upper stories infilled (structures with pilotis). Consequently, the total number of buildings studied herein is 30 different structural systems x 3 different distributions of masonry infills = 90. The three subsets of the buildings were trained separately by the same Machine Learning methods, in order to draw conclusions about the possible differences in the predictive ability of the ML techniques, resulting from the influence of the infill walls on the seismic response of the buildings. The 30 selected bare buildings (no infill walls) were modeled, analyzed and designed based on the provisions of EC2 and EC8. After the elastic modeling and design of the bare buildings, the three subsets mentioned above (bare, infilled, buildings with pilotis) were created and their nonlinear behavior was simulated, in order to analyze them by means of NTHA. Moreover, the masonry infills were modeled as single equivalent diagonal struts with stress-strain diagrams according to the model proposed by Crisafulli [50]. A detailed description and documentation of the design and modeling process of the investigated buildings can be found in [31].

A suite of 65 pairs of horizontal bidirectional earthquake records taken from the PEER [51] and the European Strong-Motion database [52] was chosen in such a way as to cover a large variety of conditions regarding tectonic environment, modified Mercalli intensity and closest distance to fault rapture, thus representing a wide range of intensities and frequency content. A detailed description and documentation of the selected earthquake records can be found in [31].

The 90 buildings (three subsets of 30 buildings each) were subjected to each one of the 65 earthquake ground motions, for which NTHA was conducted with the aid of Ruaumoko software [53]. As a consequence, a total of 5850 NTHA (90 buildings x 65 earthquake records) were conducted herein. For each one of the analyses, the estimation of the seismic damage was determined using the Maximum Interstory Drift Ratio (MIDR), which corresponds to the maximum story's drift among the perimeter frames. A detailed description and documentation of the MIDR can be found in [31]. The MIDR is extensively adopted as an reliable indicator of structural and nonstructural global damage of R/C buildings (e.g. [54-55]) and has been used by many researchers for the assessment of the building' inelastic response. The values of MIDR have been classified by many researchers. Herein, the classification given by Masi et al. [56] (Table 1) has been adopted. Note that the number of the damage classes (three) was also selected in order to be compatible with the commonly used rationale of seismic damage classification in slight (green), moderate (yellow) and heavy (red) damage states which are utilized in case of the rapid seismic assessment of buildings after strong events.

**Table 1** Relation between MIDR and damage state.

| MIDR (%) | <0.50<br>Class 0 | 0.50-1.00<br>Class 1 | >1.00<br>Class 2 |
|---|---|---|---|
| **Degree of damage** | Slight<br>(No damages or repairable slight damages) | Moderate<br>(Significant but repairable damages) | Heavy<br>(Non-repairable damages) |

## 3. Inputs and Outputs

For real problem modeling situations such as the one under consideration, the input models come from the same boundary distribution or follow a common cluster structure. Thus, the classified data enable a learning process, providing useful information for exploring the data structure of the overall set and finding patterns capable of identifying the problem, thus creating an intelligent classification framework. The classification concerns the classification of each sample in one of the predefined classes after successful training. The training of a model of machine learning with the method of classification is called the process in which the function $\hat{f}: R^N \rightarrow T$ is calculated, where T is a set of labels denoting the class. In this problem, the basic evaluation criterion was considered to be the error for a wrong prediction, which depends on the concept of the success of including a sample in the correct class.

For the purposes of the study, both structural and seismic parameters were chosen as input features in the process of the ML methods. More specifically, the following structural parameters, that are considered crucial for the vulnerability assessment of R/C buildings, were selected: the total height of buildings $H_{tot}$, the ratios of the base shear that is received by R/C walls (if they exist) along two horizontal orthogonal directions x and y (ratio $n_{vx}$ and ratio $n_{vy}$) and the structural eccentricity $e_0$. Regarding the seismic parameters, the 14 seismic parameters presented in Table 2 were chosen (e.g. [57-58]). Regarding the output feature, the abovementioned MIDR was chosen, as a reliable damage measure that can adequately capture the damage level of the R/C buildings.

**Table 2** Examined ground motion parameters

| Ground Motion Parameter | Calculation procedure | Category |
|---|---|---|
| Peak Ground Acceleration: **PGA** | max\|a(t)\| | |
| Peak Ground Velocity: **PGV** | max\|v(t)\| | |
| Peak Ground Displacement: **PGD** | max\|d(t)\| | |
| Arias Intensity: $I_a$ | $I_a = (\pi/2g) \cdot \int_0^{t_{tot}} [a(t)]^2 dt$ | Seismic parameters determined from the time histories of the records. |
| Specific Energy Density: **SED** | $SED = \int_0^{t_{tot}} [v(t)]^2 dt$ | |
| Cumulative Absolute Velocity: **CAV** | $CAV = \int_0^{t_{tot}} |a(t)| dt$ | |
| Acceleration Spectrum Intensity: **ASI** | $ASI = \int_{0.1}^{0.5} S_a(\xi = 0.05, T) dT$ | |
| Housner Intensity: **HI** | $HI = \int_{0.1}^{2.5} PSV(\xi = 0.05, T) dT$ | Seismic parameters determined from the response spectra of the records. |
| Effective Peak Acceleration: **EPA** | $EPA = (1/2.5) \{\bar{S}_a(\xi = 0.05, T)\}_{0.1}^{0.5}$ | |
| $V_{max}/A_{max}$ (**PGV/ PGA**) | max\|v(t)\|/max\|a(t)\| | Seismic parameters accounting for the earthquake's frequency content. |
| Predominant Period: **PP** | $PP = T[\max S_a(\xi = 0.05, T)]$ | |
| Time of Uniform Duration: **TUD** | | |
| Time of Bracketed Duration: **TBD** | Special algorithm<br>(e.g. SeismoSoft 2015) | Seismic parameters based on the earthquake's duration. |
| Time of Significant Duration: **TSD** | | |

Where, a(t), v(t) and d(t) are the acceleration, velocity and displacement time history respectively, $S_a$ is the acceleration spectrum,

PSV is the pseudovelocity spectrum, ξ is the damping ratio, Time of Uniform Duration is the total time during which the ground acceleration is larger than a given threshold value (usually 5% of PGA), Time of Bracketed Duration is the total time elapsed between the first and the last excursions of a specified level of acceleration (usually 5% of PGA), Time of Significant Duration is the interval of time over which a proportion of the total Arias Intensity is accumulated (usually the interval between the 5% and 95% thresholds).

## 4. Preprocessing of Data

The preprocessing of the data refers to the preliminary checks and work carried out on the abovementioned dataset before the use of the ML algorithms, in order to determine if the initial data suffer from various types of problems and if so, then to select the appropriate procedure to deal with them. This process is particularly important because, in case that the quality of the data used is not guaranteed, the performance of the ML algorithms will not be satisfactory or will produce biased or untrue results. Finally, it should be noted that there are a number of techniques which can be used in the preprocessing procedures and that the choice of the best strategy depends on the nature of the examined problem and on the corresponding available data used. In detail, the data pre-processing procedures that were applied to the present dataset include the following checks:

### 4.1 Missing Values

A Missing Values check was performed and it was found that there is no unavailable information that could mislead the algorithms and produce untrue results.

### 4.2 Outliers

An extreme value is defined as a point that is very far from the mean value of the corresponding random variable representing a feature. Samples with feature values very different from the mean value produce significant errors, especially if they are the result of noise during the measurement process, something which has disastrous results for the training process. Distance is measured relative to a threshold, which is usually a multiple of the standard deviation. For a random variable following a normal distribution, a distance that equals twice the standard deviation covers 95% of the points and a distance which equals three times the standard deviation covers 99% of the points. If the number of extreme values is small, then either the values remain and are appropriately modified or these samples are simply discarded, which is the most popular tactic.

One of the most popular methods for finding extreme values is the Interquartile Range (IQR) technique. IQR is the difference between the $3^{rd}$ (Q3) and the $1^{st}$ (Q1) quadrant, namely IQR = $Q^3 - Q^1$. Quadrants divide the data into 4 equal parts (quarters), with the intra-quadratic range comprising an intermediate 50% of observations. The remaining 50% is outside this range, with the 25% of the observations being smaller than $Q^1$ and the remaining 25% being larger than $Q^3$. A depiction of the IQR method is presented in the following Figure 1.

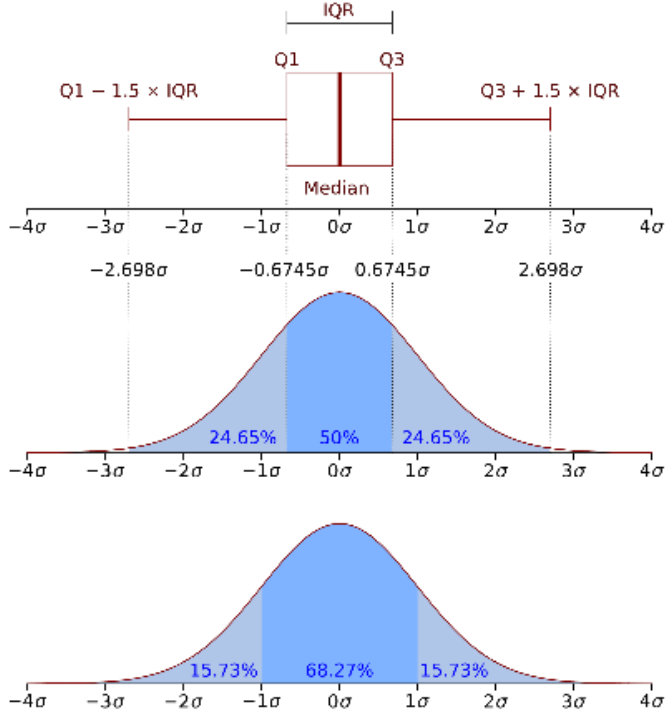
**Fig 1.** Interquartile Range

It should be emphasized that the extreme values in the case of the problem of seismic damage that is considered herein are important and consist the question of the problem, since, based on them, important decision-making mechanisms are activated (e.g., further detailed seismic evaluation of the building etc.), so it was considered appropriate to seek them, but not to isolate - remove them from the datasets. This decision was considered essential in order to create objective training samples, which will be able to generalize and to better respond to new data, as well as to be able to predict corresponding damage for future periods (forecasting).

### 4.3 Normalization

Normalization is a process of transforming data, in which numeric values are replaced by corresponding ones, but which are in a certain range of values. This process is usually performed in order to address problems related to the operation or performance of the algorithms. For example, some algorithms perform better when the input values lie in the range [0,1], while in case of algorithms that calculate the distances among the observations, the normalization of values is required in order to deal with the problem that the variables with large values are those that mostly determine the distance of the observations, while small-value variables have very little effect on the distance and, consequently, play no role in calculating the result. In the present research study, the normalization was done with the aid of the method Max-Min. According to this method, all numerical values match with values that fluctuate within a predetermined range based on a linear transformation. Considering a variable $A$, with $max_A$ and $min_A$ being the largest and smallest values respectively, we can match all the values with corresponding ones that fluctuate within a range with a lower limit of $new\_min_A$ and an upper limit of $new\_max_A$ according to the relation 1:

$$x' = \frac{x - min_A}{max_A - min_A}(new\_max_A - new\_min_A) + new\_min_A \qquad (1)$$

where x is the value of the variable $A$ and $x'$ is the new value.

This method has the advantage that the user predefines the value range, setting $new\_min_A$ and $new\_max_A$, while maintaining the ratio between the values that existed in the original data. On the other hand, the normalization of Max-Min is not appropriate in cases where the data contain extreme values, as they gather the vast majority of values in a minimal part of the value range and use the rest of the part for exceptions.

### 4.4 Feature Reduction

In most cases a set of data can contain too many features, which may be related to each other, provide irrelevant information to the specific problem or produce noise, something which reduces the efficiency of the algorithm used. Two depictions of the correlation matrix of the dataset used (regression depiction in left and heat map depiction in right) are presented in the following Fig. 2.

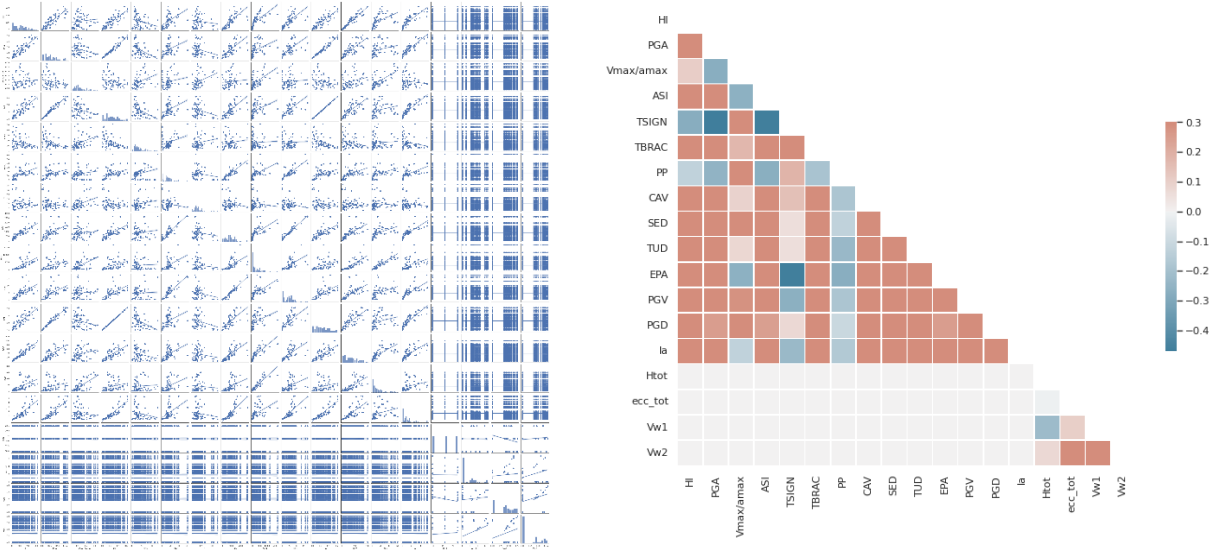

**Fig 2.** Correlation Matrix. Regression depiction (left) and heat map depiction (right)

Also, if the vector space of features has many dimensions (i.e., many features), the volume of this space increases very fast, so the data for the problem will be sparse, causing problems for the methods that try to achieve statistical significance. The amount of data needed in order to be considered dense increases exponentially in relation to the dimension of the feature space. This phenomenon is also known as the "curse of dimensionality". It should also be noted that a large number of features increases the number of parameters of the learning system, and, therefore, its complexity, without this meaning that it will have a correspondingly better performance. Because of these observations, the number of features should be kept as small as possible in order to achieve high system performance.

The solution to these problems is provided with the aid of techniques of dimensional reducing, which offer an efficient solution to managing multidimensional data, as they seek for a low-dimensional structure in multidimensional data. These techniques are considered necessary pre-processing procedures in such cases, as the distances between the data in the reduced space are calculated faster, the size of the dataset is reduced, the data structure which remains hidden in the original multidimensional space is revealed and the efficiency of ML algorithms is significantly improved. The most well-known linear dimensional reduction technique is Principal Component Analysis (PCA).

This method tries to calculate the axes in which the maximum data scatter is observed. For example, for the data $\{X_1, X_2, \ldots, X_n\} \in R^D$, the covariance table $S = X \cdot X^T$ is calculated, then their average value $\mu$ is calculated, the eigenvalues $I_i$ and the eigenvectors $e_i$ are calculated through of the process of self-analysis of $S$, $I_i \cdot e_i = S \cdot e_i$, and, finally, the $d$ largest eigenvectors are selected and based on them the new variables are calculated by the equation 2:

$$Y_i = [e_1, e_2, \ldots, e_d]^T \times (X_i - \mu) \tag{2}$$

Subsequently, a PCA test was performed for the dataset considered in the present study, in order to detect data covariance and to apply, if necessary, a feature reduction. As can be seen from the scree plot in Fig. 3, the principal components retain less than 60% of the statistical data from the original data, so no feature reduction is required.

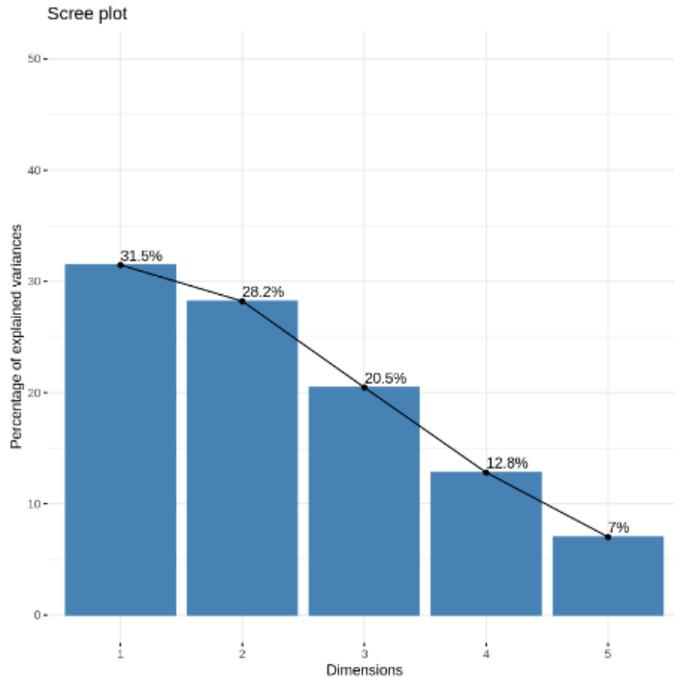

**Fig. 3**. Principle Component Analysis

### 4.5 Feature Selection

This is the process of the optimal selection of a subset of existing features without transformation, in order to retain the most important of them reducing this way their number and at the same time retaining as much useful information as possible. This step is crucial because if features with low separating ability are selected, the resulting learning system will not have satisfactory performance, while if features that provide useful information are selected, the system that will be designed will be simple and efficient.

One of the strategies that can be followed is to examine the characteristics one by one through a measure of class reparability and to reject those that have a low separating ability. The aim is to select these characteristics that lead to large distances between the groups of samples to and small variation among the same group. This means that the characteristics should get distant values for different classes and close values for the same class, a strategy known as filtering. One of the most popular filtering methods is Forward Selection, which starts with an empty set of selected features. Then, this method selects the most important of the other features, subtracts it from the original set and adds it to the set of selected features. Finally, from the remaining features, the most important one is selected and it is added to the set of selected features. The process is repeated until an output condition is satisfied.

Also, another approach to feature selection is achieved by examining the various combinations of features available and controlling those combinations that lead to higher performance, regardless of the quality of the individual features, an approach which is called wrapping. These methods, in order to select the important features, use the same algorithm that will be applied to the final ML process. In other words, these are not independent methods the results of which can be separately dealt with, but methods that differ in terms of the learning algorithm and of the solution search technique.

In the present research work, taking into account the inability of classical correlation analysis methods to detect nonlinear correlations such as sinus wave, quadratic curve, etc., the Predictive Power Score (PPS) technique was chosen to summarize the most important features between available features. PPS can work with nonlinear relationships, but also with asymmetric relationships, explaining that variable A informs variable B more than variable B informs variable A. Technically, the score is a measurement in space [0, 1] of the success of a model in predicting a variable target with the aid of an off-sample variable prediction, something which practically means that this method can increase the efficiency of finding hidden patterns in the data and selecting appropriate prediction variables. The final process of capturing the predictive power of the individual characteristics was done with the PPS technique, where for the calculation of PPS in numerical variables the metric of Mean Absolute Error (MAE) was used, which is the measure of quantification of error between estimation or prediction and the observed values. MAE is given by the equation 3:

$$MAE = \frac{1}{n}\sum_{i=1}^{n}|f_i - y_i| = \frac{1}{n}\sum_{i=1}^{n}|e_i| \qquad (3)$$

where $f_i$ is the estimated value and $y_i$ the true one. The average of the absolute value of the ratio between these values is defined as the absolute error of their relation $|e_i| = |f_i - y_i|$. The following Figs 4, 5 and 6 illustrate the predictive power of the features used in the present investigation:

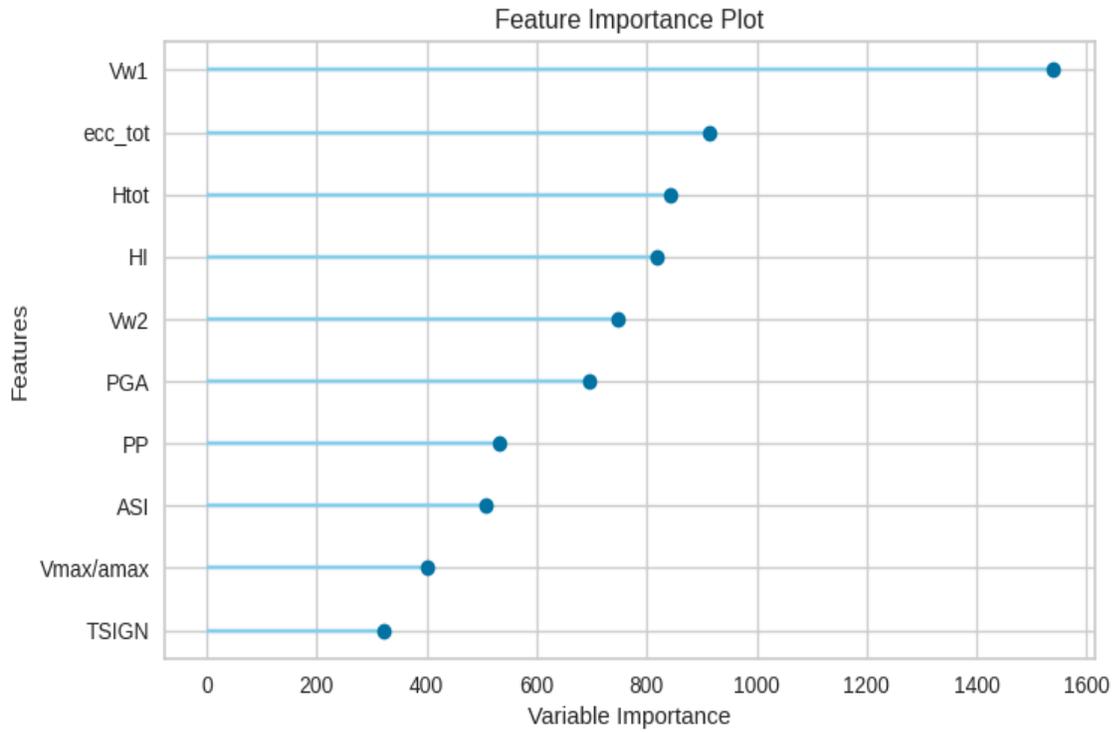

**Fig 4.** Feature Importance plot for Row_Form_Bare dataset

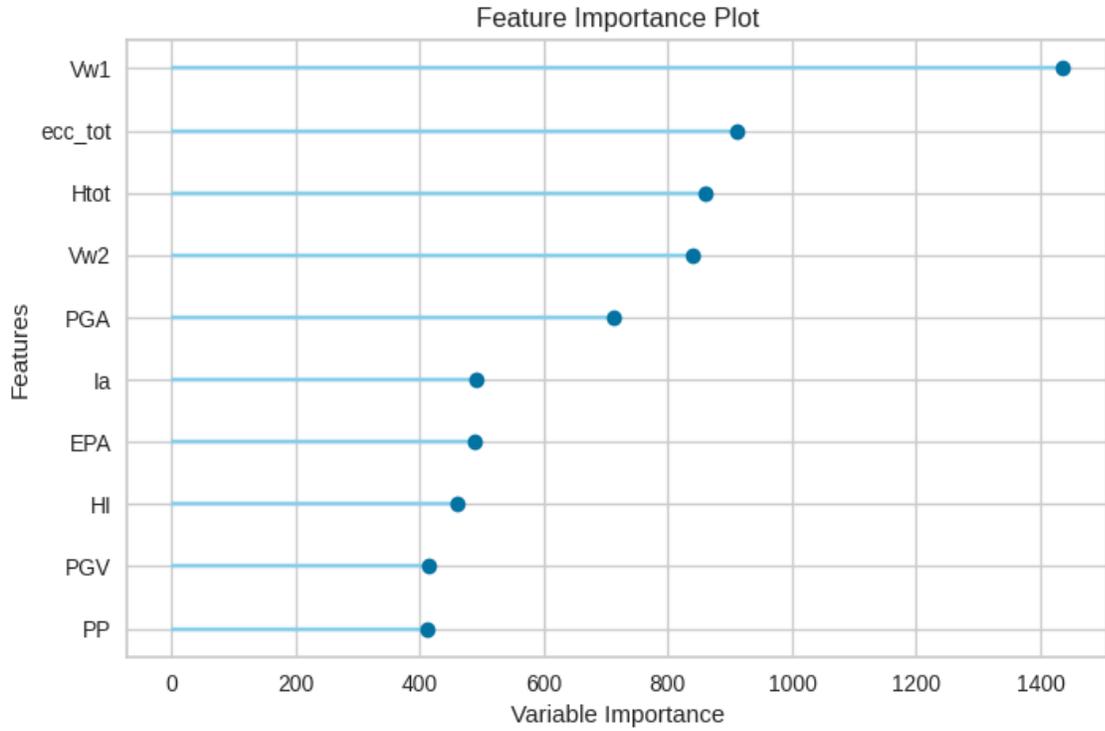

**Fig 5.** Feature Importance plot for Row_Form_Full-Masonry dataset

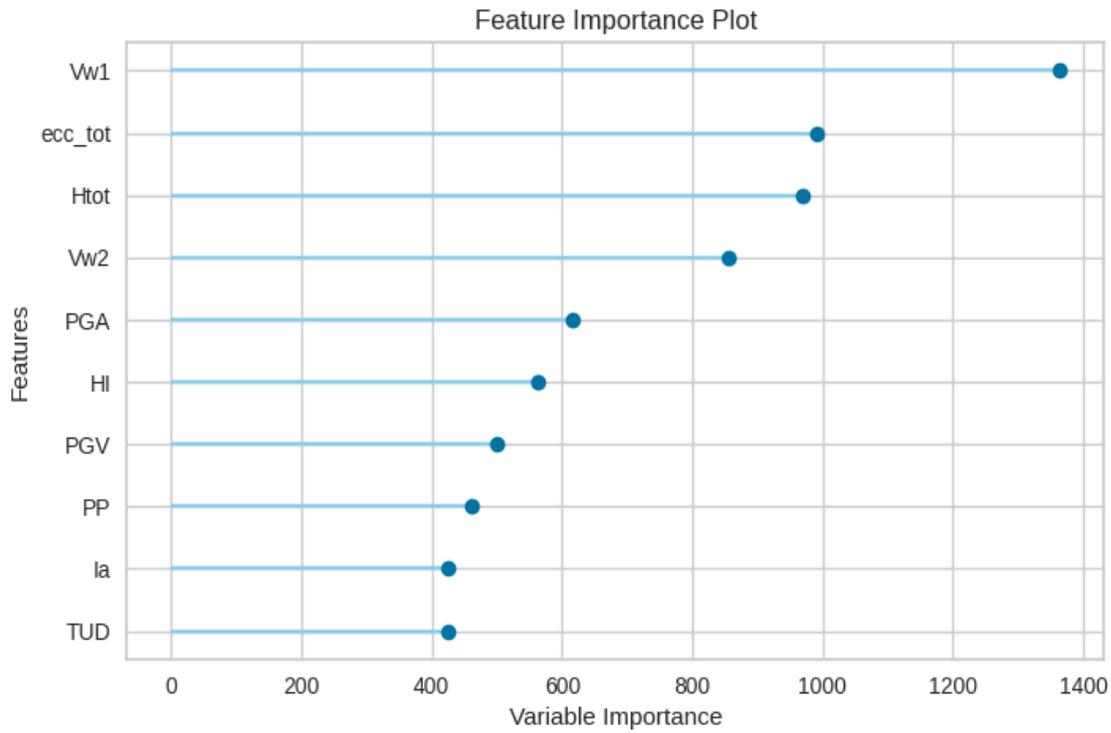

**Fig 6**. Feature Importance plot for Row_Form_Pilotis dataset

5. **Methodology**

In order to carry out a thorough investigation of the ML algorithms' ability to model the given problem based on the existing data, an extensive comparison of the most well-known algorithms was made. This comparison includes metrics for evaluating the performance of each algorithm, its execution time, generalization error, as well as the inherent behavior of the algorithm, in order to gain a deeper understanding of its theoretical basis.

### 5.1 Presentation of used machine learning algorithms

In order to identify the most effective algorithm that is capable to predict the R/C buildings' seismic damage with high accuracy, an extensive comparison with the most widely used supervised ML models was made. A comprehensive review of the comparison models is summarized as follows:

1. **Support Vector Machines (SVMs):** SVM is a supervised machine learning algorithm that can be used for both classification and regression problems. In the SVM algorithm, each data item is a point in n-dimensional space with the value of each feature being the value of a particular coordinate. The classification is achieved by finding the hyper-plane that differentiates the two classes very well [59].
2. **Random Forest Classifier:** A Random Forest is a meta-learner that builds a number of classifying decision trees on various sub-samples of the dataset and uses averaging to improve the predictive accuracy and to control over-fitting [60].
3. **CatBoost Classifier:** CatBoost is an algorithm based on gradient boosted decision trees. Gradient boosting is a machine learning technique for regression and classification problems, which produces a prediction model in the form of an ensemble of weak prediction models, typically decision trees [61].
4. **Light Gradient Boosting Machine:** Light Gradient Boosting Machine a gradient boosting framework based on decision trees to increases the efficiency of the model and reduces memory usage [62].
5. **Extreme Gradient Boosting:** This method produces an ensemble prediction model by a set of weak decision trees prediction models. It builds the model smoothly, allowing at the same time the optimization of an arbitrarily differentiable loss function [63].
6. **Extra Trees Classifier:** Extra Trees is an information-based learning methodology. Specifically, it is an ensemble machine learning algorithm that combines the predictions from many decision trees [64].
7. **Decision Tree Classifier**: A decision tree is a tree-based model including chance event outcomes and resource costs, in order to displays conditional control statements. Each node represents an attribute, each branch represents the outcome of an attribute test and each leaf represents the decision taken after computing all attributes. The paths from the root to leaf represent the regression process [65].
8. **Gaussian Process Classifier:** The Gaussian process is a stochastic process (a collection of random variables indexed by time or space), such that every finite collection of those random variables has a multivariate normal distribution, i.e. every finite linear combination of them is normally distributed. Gaussian Process classifier is a generalization of the Gaussian probability distribution and can be used as the basis for sophisticated non-parametric machine learning algorithms for classification and regression [66].
9. **k-Neighbors Classifier:** k-Nearest Neighbors is a similarity-based learning algorithm, according to which the target is predicted by local interpolation of the targets associated with the nearest neighbors in the training set [67].
10. **Linear Discriminant Analysis (LDA):** LDA is a generalization of Fisher's linear discriminant, a method used in statistics and other fields to find a linear combination of features that characterizes or separates two or more classes of objects or events. The resulting combination may be used as a linear classifier, or more commonly, for dimensionality reduction before later classification [68].
11. **Ridge Classifier:** Ridge Regression is a regression method that does not provide confidence limits. It uses regularization L2-norm in order to solve a high covariance problem, even if the errors come from an abnormal distribution [69].
12. **Quadratic Discriminant Analysis (QDA):** QDA is a generative model that assumes that each class follows a Gaussian distribution. The class-specific prior is simply the proportion of data points that belong to the class. The class-specific mean vector is the average of the input variables that belongs to the class [70].
13. **MLP Classifier:** Multi-layer Perceptron (MLP) is a supervised learning algorithm that trains using Backpropagation. It can learn a non-linear function approximator for either classification or regression. It is different from logistic regression in that between the input and the output layer there can be one or more non-linear layers, called hidden layers [71].

14. **Naive Bayes:** Naive Bayes methods are a set of supervised learning algorithms based on applying Bayes' theorem with the "naive" assumption of conditional independence between every pair of features given the value of the class variable [72].
15. **AdaBoost Classifier:** It is a meta-learner that begins by fitting a regressor on the original dataset and then fits additional copies of the regressor on the same dataset where the weights of instances are adjusted according to the error of the current prediction [73].
16. **Logistic Classifier:** The logistic classifier model is a classification model in which the conditional probability of one of the two possible realizations of the output variable is assumed to be equal to a linear combination of the input variables, transformed by the logistic function [74].

### 5.2 Data Sampling

In order to have an objective evaluation process of ML models, both as a way of self-evaluation and for their comparison with the corresponding alternative models, there are various statistical techniques of distribution and handling of datasets, which are also called validation techniques. K-Fold is the most common cross-validation method, according to which the dataset is randomly divided into k subsets, each of relatively equal population. Of the aforementioned k subsets, one is used as a test subset, while the all-theoretic compound of the remaining k-1 subsets is used as a training subset. A total of k computing cycles are performed, so that, in turn, each of the k subsets is used as a test subset. The advantage of this evaluation method is that each data is used for training and definitely once for examination. The parameter *k* can attain any positive integer value, while the most popular choice in practical applications is the case where *k*=10, which is called 10-Fold Cross Validation. A depiction of the 10-Fold Cross Validation method is presented in the following Fig. 7.

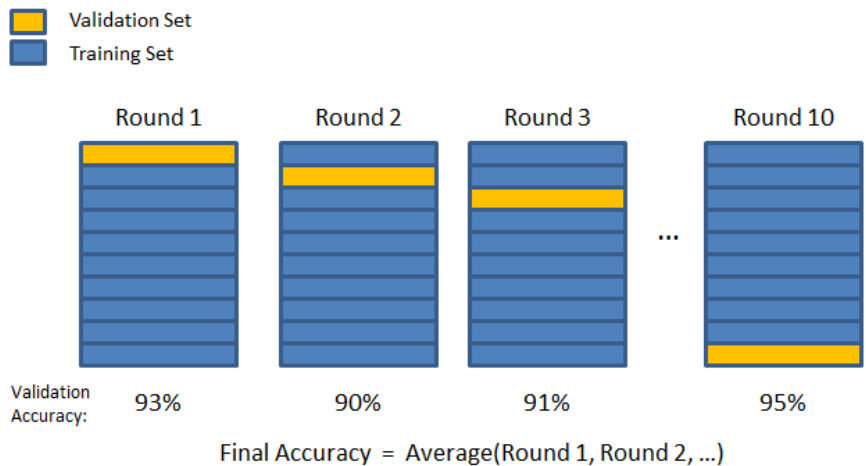

**Fig 7.** 10-Fold Cross Validation

### 5.3 Comparison Results

For the thorough evaluation of the problem of buildings' seismic damage classification in the respective categories of damage level, a thorough investigation study was carried out with various machine ML. We should mention that the following comparison is based on different performance metrics that are used to evaluate the different machine learning algorithms for the specific classification problem. Specifically, the classification performance metrics which used are Accuracy, Receiver Operating Characteristic (ROC), Recall, Precision, F-Score, Cohen's Kappa Statistics (CKS), and Matthews Correlation Coefficient (MCC). All metrics are the simple ratios between the number of correctly classified points to the total number of points, in the test dataset (unseen data points). The best performance is achieved by the model with a value near 1. An explanation of the classification metrices is depicted in the following Fig. 8.

**Fig 8.** Classification metrices explained

The results by descent performance (accuracy) order for each dataset are presented in the following Tables 3, 4 and 5.

Table 3. Performance Metrics in the Row_Form_Bare dataset

| ID | Model | Accuracy | ROC | Recall | Precision | F-Score | CKS | MCC | Time/sec |
|---|---|---|---|---|---|---|---|---|---|
| 1. | SVM - Gaussian Kernel | 0.8849 | 0.9739 | 0.8683 | 0.8844 | 0.8843 | 0.8205 | 0.8208 | 0.659 |
| 2. | Random Forest Classifier | 0.8772 | 0.9691 | 0.8578 | 0.8756 | 0.8758 | 0.8081 | 0.8087 | 0.567 |
| 3. | CatBoost Classifier | 0.8757 | 0.9747 | 0.8553 | 0.8744 | 0.8744 | 0.8056 | 0.8063 | 4.158 |
| 4. | Light Gradient Boosting Machine | 0.8664 | 0.9724 | 0.8467 | 0.8659 | 0.8657 | 0.7915 | 0.7920 | 0.246 |
| 5. | Extreme Gradient Boosting | 0.8649 | 0.9716 | 0.8455 | 0.8654 | 0.8645 | 0.7892 | 0.7898 | 6.954 |
| 6. | Extra Trees Classifier | 0.8633 | 0.9635 | 0.8450 | 0.8630 | 0.8626 | 0.7868 | 0.7873 | 0.526 |
| 7. | Decision Tree Classifier | 0.8548 | 0.8915 | 0.8388 | 0.8573 | 0.8552 | 0.7744 | 0.7751 | 0.022 |
| 8. | SVM - RBF Kernel | 0.8471 | 0.9384 | 0.8228 | 0.8467 | 0.8449 | 0.7604 | 0.7623 | 0.359 |
| 9. | Gaussian Process Classifier | 0.8402 | 0.9130 | 0.8192 | 0.8407 | 0.8395 | 0.7508 | 0.7517 | 2.553 |
| 10. | k-Neighbors Classifier | 0.8224 | 0.9366 | 0.7992 | 0.8228 | 0.8213 | 0.7232 | 0.7245 | 0.120 |
| 11. | Linear Discriminant Analysis | 0.8124 | 0.9479 | 0.7920 | 0.8172 | 0.8134 | 0.7098 | 0.7111 | 0.021 |
| 12. | SVM - Polynomial Kernel | 0.8008 | 0.9308 | 0.7718 | 0.7968 | 0.7981 | 0.6885 | 0.6892 | 1.002 |
| 13. | Ridge Classifier | 0.7985 | 0.0000 | 0.7452 | 0.7859 | 0.7773 | 0.6785 | 0.6895 | 0.020 |
| 14. | Quadratic Discriminant Analysis | 0.7923 | 0.9419 | 0.7928 | 0.8186 | 0.7968 | 0.6854 | 0.6933 | 0.022 |
| 15. | MLP Classifier | 0.7483 | 0.9023 | 0.7383 | 0.7781 | 0.7486 | 0.6158 | 0.6277 | 0.320 |
| 16. | Naive Bayes | 0.7320 | 0.9205 | 0.7387 | 0.7820 | 0.7421 | 0.5992 | 0.6124 | 0.020 |

| ID | Model | Accuracy | ROC | Recall | Precision | F-Score | CKS | MCC | Time/sec |
|---|---|---|---|---|---|---|---|---|---|
| **17.** | Ada Boost Classifier | 0.6797 | 0.8178 | 0.6552 | 0.7575 | 0.6826 | 0.5124 | 0.5369 | 0.146 |
| **18.** | Logistic Classifier | 0.6564 | 0.0000 | 0.5728 | 0.6130 | 0.5975 | 0.4246 | 0.4783 | 0.064 |

**Table 4**. Performance Metrics in the Row_Form_Full-Masonry dataset

| ID | Model | Accuracy | ROC | Recall | Precision | F-Score | CKS | MCC | Time/sec |
|---|---|---|---|---|---|---|---|---|---|
| **1.** | SVM - Gaussian Kernel | 0.8949 | 0.9777 | 0.8770 | 0.8970 | 0.8941 | 0.8197 | 0.8218 | 0.244 |
| **2.** | Extreme Gradient Boosting | 0.8942 | 0.9759 | 0.8745 | 0.8976 | 0.8935 | 0.8187 | 0.8211 | 15.896 |
| **3.** | CatBoost Classifier | 0.8926 | 0.9763 | 0.8710 | 0.8950 | 0.8921 | 0.8154 | 0.8172 | 4.328 |
| **4.** | Random Forest Classifier | 0.8918 | 0.9739 | 0.8685 | 0.8961 | 0.8918 | 0.8145 | 0.8169 | 0.562 |
| **5.** | Light Gradient Boosting Machine | 0.8864 | 0.9747 | 0.8635 | 0.8914 | 0.8861 | 0.8053 | 0.8082 | 0.665 |
| **6.** | Extra Trees Classifier | 0.8834 | 0.9690 | 0.8577 | 0.8860 | 0.8830 | 0.7998 | 0.8018 | 0.516 |
| **7.** | Decision Tree Classifier | 0.8749 | 0.8986 | 0.8550 | 0.8779 | 0.8743 | 0.7865 | 0.7887 | 0.021 |
| **8.** | SVM - RBF Kernel | 0.8726 | 0.9498 | 0.8388 | 0.8765 | 0.8716 | 0.7806 | 0.7832 | 0.387 |
| **9.** | K Neighbors Classifier | 0.8687 | 0.9494 | 0.8336 | 0.8700 | 0.8666 | 0.7727 | 0.7753 | 0.128 |
| **10.** | Gaussian Process Classifier | 0.8656 | 0.9290 | 0.8358 | 0.8705 | 0.8658 | 0.7707 | 0.7731 | 2.605 |
| **11.** | Linear Discriminant Analysis | 0.8324 | 0.9442 | 0.7966 | 0.8390 | 0.8340 | 0.7134 | 0.7150 | 0.023 |
| **12.** | Quadratic Discriminant Analysis | 0.8184 | 0.9365 | 0.8051 | 0.8350 | 0.8235 | 0.6957 | 0.6994 | 0.021 |
| **13.** | SVM - Polynomial Kernel | 0.8015 | 0.0000 | 0.7338 | 0.7896 | 0.7840 | 0.6453 | 0.6568 | 0.019 |
| **14.** | Ada Boost Classifier | 0.7837 | 0.8572 | 0.7465 | 0.8123 | 0.7898 | 0.6342 | 0.6411 | 0.149 |
| **15.** | MLP Classifier | 0.7443 | 0.8879 | 0.6845 | 0.7293 | 0.7322 | 0.5531 | 0.5577 | 0.988 |
| **16.** | Naive Bayes | 0.7405 | 0.9160 | 0.7636 | 0.7944 | 0.7525 | 0.5876 | 0.6040 | 0.020 |
| **17.** | Logistic Classifier | 0.6988 | 0.8728 | 0.6076 | 0.6877 | 0.6541 | 0.4719 | 0.5056 | 0.322 |
| **18.** | Ridge Classifier | 0.5476 | 0.0000 | 0.4867 | 0.5936 | 0.5150 | 0.2506 | 0.2980 | 0.067 |

**Table 5**. Performance Metrics in the Row_Form_Pilotis dataset

| ID | Model | Accuracy | ROC | Recall | Precision | F-Score | CKS | MCC | Time/sec |
|---|---|---|---|---|---|---|---|---|---|
| **1.** | SVM - Gaussian Kernel | 0.8795 | 0.9744 | 0.8518 | 0.8823 | 0.8792 | 0.8109 | 0.8125 | 6.119 |
| **2.** | Light Gradient Boosting Machine | 0.8772 | 0.9754 | 0.8456 | 0.8776 | 0.8759 | 0.8067 | 0.8080 | 0.246 |
| **3.** | CatBoost Classifier | 0.8772 | 0.9749 | 0.8433 | 0.8776 | 0.8752 | 0.8064 | 0.8082 | 4.387 |
| **4.** | Extreme Gradient Boosting | 0.8664 | 0.9710 | 0.8318 | 0.8673 | 0.8643 | 0.7895 | 0.7916 | 0.662 |

| ID | Model | Accuracy | ROC | Recall | Precision | F-Score | CKS | MCC | Time/sec |
|---|---|---|---|---|---|---|---|---|---|
| 5. | Random Forest Classifier | 0.8626 | 0.9677 | 0.8285 | 0.8658 | 0.8623 | 0.7841 | 0.7857 | 0.561 |
| 6. | Extra Trees Classifier | 0.8479 | 0.9622 | 0.8081 | 0.8505 | 0.8472 | 0.7607 | 0.7626 | 0.512 |
| 7. | Decision Tree Classifier | 0.8402 | 0.8820 | 0.8040 | 0.8432 | 0.8404 | 0.7493 | 0.7505 | 0.022 |
| 8. | SVM - RBF Kernel | 0.8370 | 0.9276 | 0.7900 | 0.8354 | 0.8326 | 0.7406 | 0.7440 | 0.363 |
| 9. | Gaussian Process Classifier | 0.8216 | 0.8974 | 0.7863 | 0.8269 | 0.8228 | 0.7207 | 0.7221 | 2.571 |
| 10. | k-Neighbors Classifier | 0.8162 | 0.9280 | 0.7772 | 0.8208 | 0.8161 | 0.7114 | 0.7135 | 0.122 |
| 11. | Ada Boost Classifier | 0.8162 | 0.8382 | 0.7897 | 0.8323 | 0.8213 | 0.7150 | 0.7179 | 0.146 |
| 12. | SVM - Polynomial Kernel | 0.8061 | 0.9375 | 0.7539 | 0.7991 | 0.8008 | 0.6927 | 0.6944 | 0.971 |
| 13. | MLP Classifier | 0.8054 | 0.9495 | 0.7777 | 0.8212 | 0.8111 | 0.6987 | 0.7008 | 0.024 |
| 14. | Ridge Classifier | 0.7976 | 0.0000 | 0.7111 | 0.7769 | 0.7710 | 0.6705 | 0.6826 | 0.020 |
| 15. | Quadratic Discriminant Analysis | 0.7807 | 0.9413 | 0.7845 | 0.8412 | 0.7962 | 0.6704 | 0.6851 | 0.020 |
| 16. | Logistic Classifier | 0.7374 | 0.9003 | 0.6757 | 0.7679 | 0.7126 | 0.5884 | 0.6144 | 0.229 |
| 17. | Naive Bayes | 0.7305 | 0.9251 | 0.7327 | 0.8013 | 0.7470 | 0.5977 | 0.6156 | 0.020 |
| 18. | Linear Discriminant Analysis | 0.6502 | 0.0000 | 0.5548 | 0.6278 | 0.5930 | 0.4302 | 0.4812 | 0.064 |

### 5.4 Best Performance Algorithm

In all three cases examined the Support Vector Machine (SVM) - Gaussian Kernel algorithm produced the highest classification results. The basic function of SVMs is to construct a super-level that plays the role of a decision-making surface, so that the margin of separation of the categories is maximized. A key feature of SVMs that determines their function is the so-called support vectors, which consist a small subset of the training data used. Considering the problem of categorizing two categories, as described at one level in Fig. 9 (left), it is obvious that the two categories marked with the labels "+" and "o" are linearly separable. However, there are many lines $\varepsilon_1, \varepsilon_2, \varepsilon_3, ...$ which are multiple possible decision surfaces that can achieve the same result. The SVM algorithm seeks for the single line ($\varepsilon^*$) that separates the categories in such a way that the margin between the categories is maximized, as shown in Fig. 9(right), and consists the optimal decision surface.

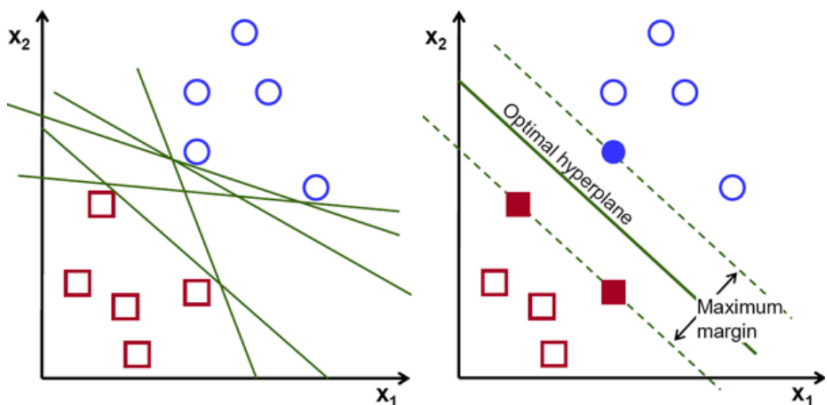

**Fig 9.** Problem of two linearly separable categories

In this case the data are linearly separable, something which guarantees the error-free classification of the data. As a consequence, the problem is reduced to the simplest case of patterns' classification, since the decision-making surface has the following simple form:

$$w^T x + b = 0 \tag{4}$$

where $x$ is the input vector, $w$ is the vector of the weights and $b$ is the bias constant to be calculated. Because the data are linearly separable, the categorizer is described by the following equations 5 and 6:

$$w^T x_k + b \geq +1, for\ t_k = +1 \tag{5}$$

or

$$w^T x_k + b \leq -1, for\ t_k = -1 \tag{6}$$

The above two equations can be described together, using the equation 7:

$$t_k(w^T x_k + b) \geq +1, \quad k = 1,2,3,\ldots,N \tag{7}$$

The goal of SVMs is to find the decision-making surface by maximizing the margin that separates the categories, which equals to $\frac{2}{\|w\|_2}$. The vectors for which the equality of the above function applies are the so-called support vectors and are those vectors that lie closest to the decision-making surface and, therefore, those that are more difficult to categorize than all training vectors. Therefore, the problem of classification becomes an optimization problem, in which the optimal surface $(w^*, b^*)$ that reduces the cost $J(w) = \frac{1}{2} w^T w$ satisfying some constraints is searched. This problem is defined as follows by the equation 8:

$$\min_{w,b} \left\{ J(w) = \frac{1}{2} w^T w \right\} \tag{8}$$

so that $t_k(w^T x_k + b) \geq +1,\ k = 1,2,3,\ldots,N$.

In the above optimization problem, which is called primal, the cost function is convex and the constraints are linear with respect to $w$. The solution is achieved with the aid of the Lagrange multipliers method, based on which the following Lagrange function is formed by equation 8:

$$L(w, b, a) = \frac{1}{2} w^T w - \sum_{k=1}^{N} a_k [t_k(w^T x_k + b) - 1] \tag{9}$$

where the coefficients $a_k \geq 0, k = 1, \ldots, N$ are called Lagrange multipliers.

The solution of the initial optimization problem with constraints becomes an $L(w,b,a)$ saddle point optimization problem. In particular, this point should be maximized with respect to $a$ and minimized with respect to $w$ and $b$, by equation 10:

$$\max_{a} \min_{w,b} L(w, b, a) \tag{10}$$

Taking the derivatives of the function and setting them equal to zero, the following two equations 11 and 12 arise:

$$\frac{\partial L(w, b, a)}{\partial w} = 0 \tag{11}$$

$$\frac{\partial L(w, b, a)}{\partial b} = 0 \tag{12}$$

From the two conditions of the function the following equations 13 and 14 of a sigma point are derived:

$$w = \sum_{k=1}^{N} a_k t_k x_k \tag{13}$$

$$w = \sum_{k=1}^{N} a_k t_k = 0 \qquad (14)$$

Substituting the above value of $w$ into the function the dual optimization problem results, which is defined as follows by equations 15 and 16:

$$\min_{a} Q(a) = \sum_{k=1}^{N} a_k - \frac{1}{2} \sum_{l=1}^{N} \sum_{m=1}^{N} a_l a_m t_l t_m x_l^T x_m \qquad (15)$$

$$\sum_{k=1}^{N} a_k t_k = 0 \ \mu\varepsilon\ a_k \geq 0, k = 1, \dots, N \qquad (16)$$

The above becomes a Quadratic Programming problem, resulting in several non-zero $a_k$ solutions which are the requested support vectors.

By finding the optimal Lagrange multipliers $a_k^*$, the weights $w^*$ are calculated, while the corresponding bias $b^*$ is determined from one of the data separation cases. In the opposite case (which is the most probable as most problems are non-linearly separable due to uncertainty, inaccuracy of representation and noise) there is a classification error, so the purpose of SVMs is to minimize this error. For this purpose, a new set of positive numbers called slack parameters is introduced, which measure the deviation of the data from the correct classification. In this case, the decision-making surface has the form of equation 17:

$$t_k(w^T x_k + b) \geq 1 - \xi_k, \qquad k = 1,2,3, \dots, N \qquad (17)$$

where $\xi_k \geq 0$ are the slack parameters, while the corresponding initial function optimization problem is transformed in the equation 18 as follows:

$$\min_{w,b} \left\{ J(w, \xi) = \frac{1}{2} w^T w + c \sum_{k=1}^{N} \xi_k \right\} \qquad (18)$$

so that $t_k(w^T x_k + b) \geq 1 - \xi_k$, $\xi_k \geq 0$, $k = 1,2,3, \dots, N$ where $c$ is a positive constant which is usually determined experimentally. The corresponding Lagrange equation 19 will take the form:

$$L(w, b, \xi, a) = \frac{1}{2} w^T w + c \sum_{k=1}^{N} \xi_k - \sum_{k=1}^{N} a_k [t_k(w^T x_k + b) - 1 + \xi_k] - \sum_{k=1}^{N} v_k \xi_k \qquad (19)$$

where $v_k \geq 0$, $k = 1, \dots, N$ is a second (in addition to $\alpha_k$) set of Lagrange multipliers.

In this case the sagmatics point optimization problem using slack parameters is described by equation 20 as follows:

$$\max_{a,v} \min_{w,b,\xi} L(w, b, \xi, a, v) \qquad (20)$$

Finally, the problem of Quadratic Programming with slack parameters is defined by following equations 21 and 22 as follows:

$$\min_{a} Q(a) = \sum_{k=1}^{N} a_k - \frac{1}{2} \sum_{l=1}^{N} \sum_{m=1}^{N} a_l a_m t_l t_m x_l^T x_m \qquad (21)$$

$$\sum_{k=1}^{N} a_k t_k = 0, 0 \leq a_k \leq c, k = 1, \dots, N \qquad (22)$$

with the additional restriction $a_k \leq c$. The bias $b^*$ is calculated for those $a_k \leq c$ for which $\xi_k = 0$.

A major boost to the implementation of real problems was the development of nonlinear SVMs, which are based on the assumption that a nonlinearly separable pattern recognition problem can be transformed into a linearly separable one in a multidimensional space. The transformation from a space with few dimensions (input space) to a multi-dimensional space (feature space) can be achieved by applying a non-linear mapping $\varphi(x)$. In this case, the decision-making surface is defined by equation 23 as follows:

$$\sum_{i=1}^{m} w_i \varphi_i(x) + b = 0 \tag{23}$$

where $m$ is the dimension of the whole set of the nonlinear transformations $\varphi(x)$, i.e. the dimension of the feature space, which is typically much larger than the dimension $n$ of the input space. Assuming that $\varphi_0(x) = 1, \forall x, w_0 = b$ και $\varphi(x) = [\varphi_0(x), \varphi_1(x), \dots, \varphi_m(x)]^T$, the function can be written with the following form of the equation 24:

$$\sum_{i=1}^{m} w_i \varphi_i(x) = w^T \varphi(x) = 0 \tag{24}$$

Considering that by using the mapping functions $\varphi(x)$ the problem has been reduced to a linear one with separable data in the space of the features, the solution of the Lagrange function for the whole set of the weights takes the form of the equation 25:

$$w = \sum_{k=1}^{N} a_k t_k \, \varphi(x_k) \tag{25}$$

And so, it is transformed into equation 26:

$$\sum_{k=1}^{N} a_k t_k \, \varphi^T(x_k) \varphi(x) = 0 \tag{26}$$

The quantity $\varphi^T(x_k)\varphi(x)$ describes the interior product of two vectors in the feature space. This quantity is called the kernel and is denoted by equation 27:

$$K(x_k, x) = \varphi^T(x_k)\varphi(x) \tag{27}$$

Based on Mercer's theorem, the kernel can be represented by equation 28 as:

$$K(x_k, x) = \sum_{i=0}^{m} \varphi_i(x_k)\varphi_i(x), k = 1, 2, \dots, N \tag{28}$$

a technique which is called kernel trick. A depiction of the kernel trick method is represented in the following Fig. 10.

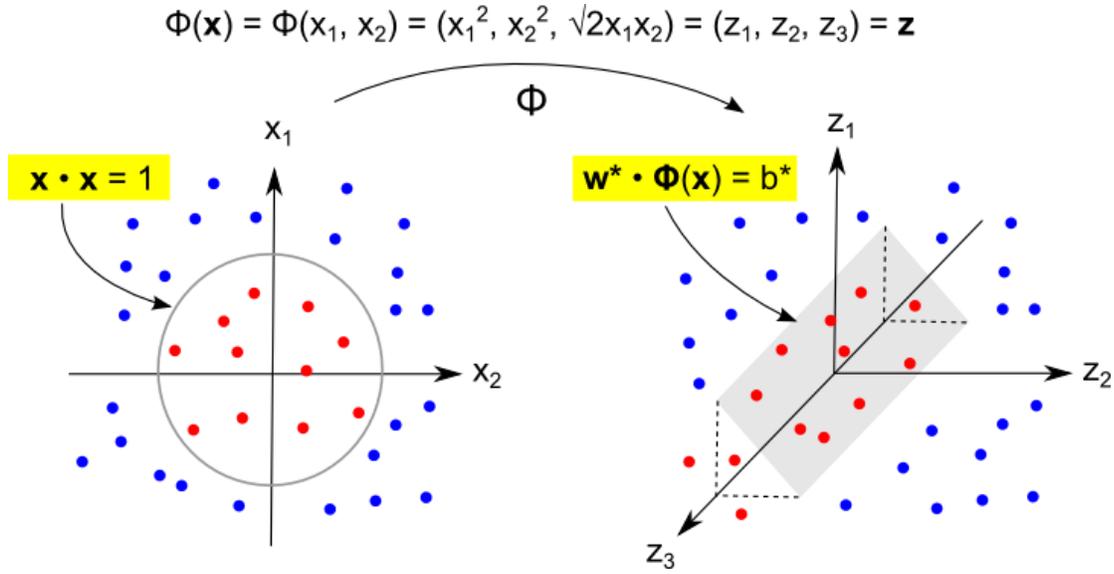

**Fig 10.** Kernel Trick

Therefore, the decision-making surface will have the form of equation 29:

$$\sum_{k=1}^{N} a_k t_k K(x_k, x) = 0 \qquad (29)$$

The corresponding dual quadratic programming optimization problem is defined by equations 30 and 31 as follows:

$$\min_{a} Q(a) = \sum_{k=1}^{N} a_k - \frac{1}{2} \sum_{l=1}^{N} \sum_{m=1}^{N} a_l a_m t_l t_m K(x_l, x_m) \qquad (30)$$

$$\sum_{k=1}^{N} a_k t_k = 0, a_k \geq 0, k = 1, \dots, N \qquad (31)$$

By finding the Lagrange multipliers from the above optimization problem the set of optimal weights $w^*$ is calculated by the equation 32:

$$w^* = \sum_{k=1}^{N} a_k^* t_k \, \varphi(x_k) \qquad (32)$$

where the first weight of the vector $w^*$ corresponds to the optimal bias $b^*$.
It is worth noting that the selection of the appropriate kernel plays an important role in the performance of SVM. The only limitations that a kernel should satisfy are that the kernel must be symmetric.
The kernels that were used in the present investigation are the following (equations 33, 34 and 35):

$$K_{polynomial}(x_k, x) = (\tau + x_k^T x)^d \qquad (33)$$

$$K_{RBF}(x_k, x) = exp\left(-\frac{1}{2\sigma^2} \|x - x_k\|_2^2\right) \qquad (34)$$

$$K_{gaussian}(x_k, x) = exp(-\gamma \|u - v\|2) \qquad (35)$$

## 5.5 Results and discussion

When building and optimizing a classification model, measuring how accurately it predicts the expected outcome is crucial. However, one metric alone can offer misleading results. There are several performance evaluations metrics to help tease out more meaning in a model. The metrics to evaluate a machine learning model are very important as the choice of metrics influences how the performance of machine learning algorithms can be compared.

Appropriate classification metrics were used in order to confirm the comparison results and to demonstrate the superiority of the machine learning algorithm that achieved the highest performance during the classification process. Specifically, given a pair of training vectors $\{x_i, y_i\}$, a classification model learns the parameters $\theta$ for an unknown function $f(x)$, which can match each input vector $x_i$ to the estimated output $f(x_i)$. Successful training means the optimal adaptation of the internal parameters $\theta$, in order to minimize an error or cost function, which evaluates the performance of the categorizer based on some efficient and sound measures. The most popular evaluation measures, which are able to evaluate and compare with clarity, completeness and objectivity the classification algorithms are presented below:

### 5.5.1 Confusion Matrix

Because incorrect classifications of different classes have different costs, it is important to assess the predictor's ability to predict each class. To evaluate the performance for each class, the following terminology is used:
1. Positive: The observations which belong to a value of the class.
2. Negative: The observations which belong to the other value of the class.
3. True Positive (TP): The number of successful predictions for positive observations.
4. True Negative (TN): The number of successful predictions for negative observations.
5. False Positive (FP): The number of failed predictions for negative observations.
6. False Negative (FN): The number of failed predictions for positive observations.

The evaluation of a classification model is based on the number of records in the control set that are correctly or incorrectly predicted by the model. This number is placed in a confusion matrix, which is a two-dimensional table, where the columns correspond to the predictions and the rows correspond to the actual values of the class.

**Table 6**. Confusion Matrix

|  |  | Predicted Class | |
|---|---|---|---|
|  |  | Class 1 | Class 0 |
| True Class | Class 1 | $f_{11}$ TP | $f_{10}$ FN |
|  | Class 0 | $f_{01}$ FP | $f_{00}$ TN |

The principle of the Confusion Matrix is that it recognizes the nature of the errors, as well as their quantity. Each snapshot $f_{ij}$ shows the number of records from class $i$ that are expected to belong to class $j$. The $f_{01}$ snapshot is the number of records from class 0 that were incorrectly predicted to be placed in class 1. Based on the snapshots, the number of records correctly predicted is the sum of $f_{00}$ και $f_{11}$, while those predicted incorrectly are $f_{01}$ and $f_{10}$. Although the Confusion Matrix provides the exact information needed to evaluate a model, this information can be expressed with aid of a unique number that is easy to use for comparisons between different models. Most performance measures can be expressed in relation to the number of TP, TN, FP and FN classifications for each class.

In the following Figs 11, 12 and 13 are presented the confusion matrices of the seismic datasets by the SVM - Gaussian Kernel algorithm which produced the highest classification results. The Confusion Matrices visualizes the prediction score that takes a fitted classifier and a set of test X and y values and returns a report showing how each of the test values predicted classes compare to their actual classes. We use confusion matrices to understand which classes are most easily confused. Also, they provide deeper insight into the classification of individual data points.

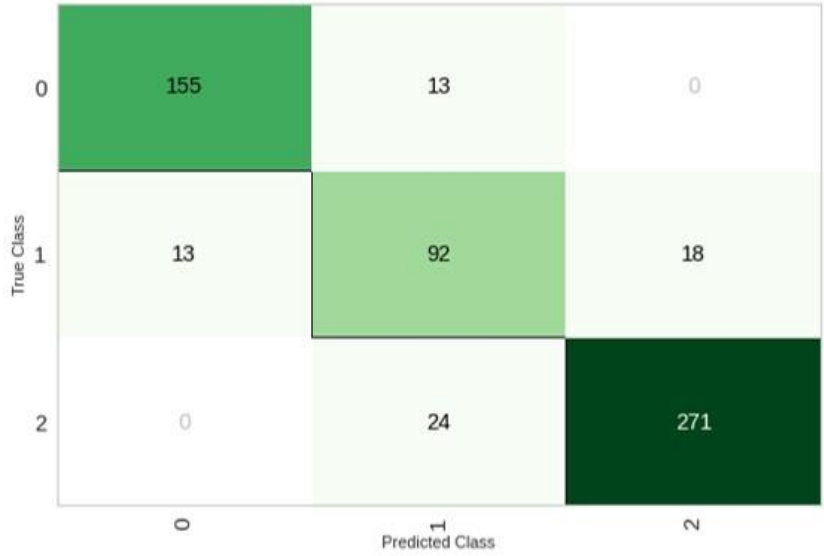
**Fig 11.** Confusion Matrix of the Row_Form_Bare dataset (for classes' definition see Table 1)

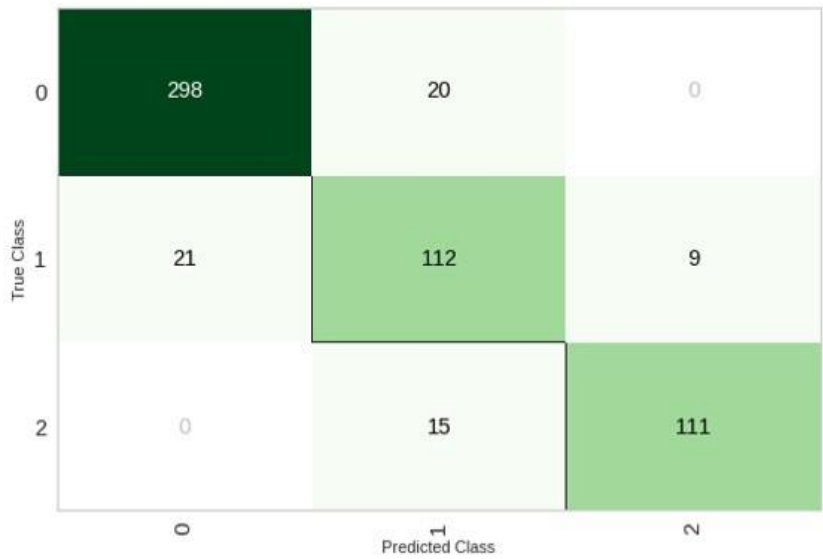
**Fig 12.** Performance Metrics of the Row_Form_Full-Masonry dataset (for classes' definition see Table 1)

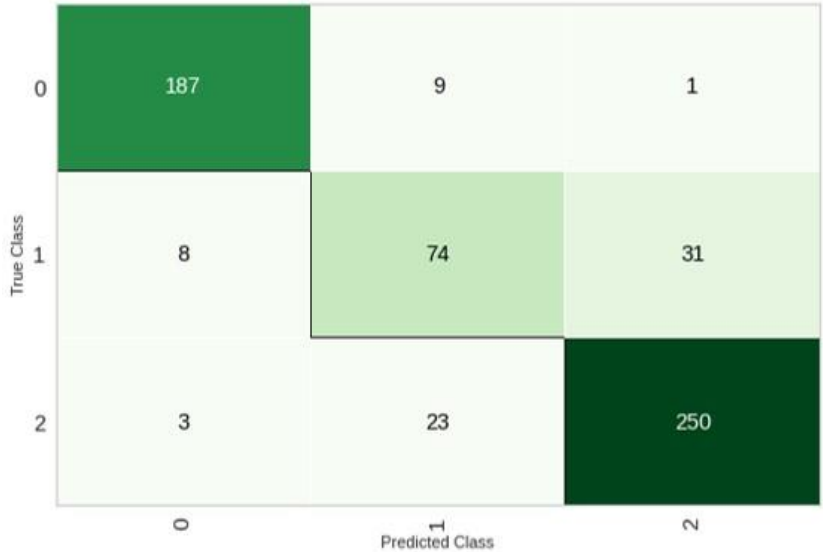

**Fig 13.** Performance Metrics of the Row_Form_Pilotis dataset (for classes' definition see Table 1)

The above confusion matrices show each combination of the true and predicted classes for each test data set. The true mode is selected, 100% accurate predictions are highlighted in green. Is the fact that the line from the True Positives which are at the top-left corner to True Negatives which are at the down-right corner, includes in the three confusion matrices most of the correct predictions in comparison to the actual values of the class.

### 5.5.2 Accuracy
It is calculated by the following relation 35:

$$accuracy = \frac{TP + TN}{TP + TN + FP + FN} \tag{35}$$

and expresses the percentage of classification of control plots that are correctly categorized.

### 5.5.3 Precision
It is calculated by the following relation 36:

$$precision = \frac{TP}{TP + FP} \tag{36}$$

and expresses the percentage of classification of the positive results that the categorizer has correctly classified as positive and are indeed positive. The higher the percentage of precision, the lower the corresponding percentage of FP.

The following Fig. 14, is a depiction of the purpose of the precision metrics.

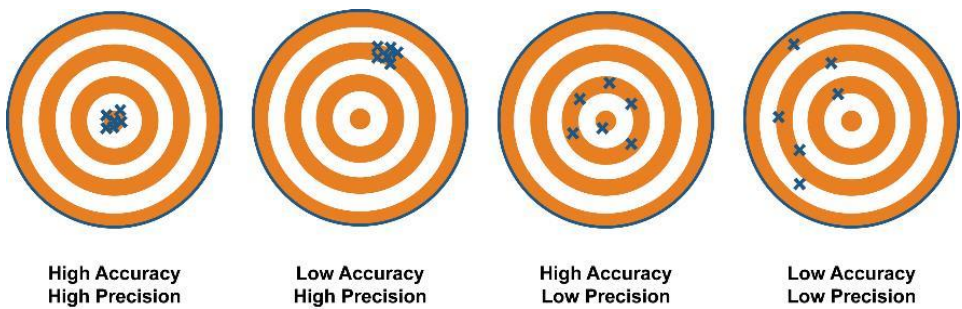

**Fig 14.** Accuracy vs Precision

### 5.5.4 Recall
It is calculated by the following relation 37:

$$recall = \frac{TP}{TP + FN} \qquad (37)$$

and expresses the percentage of classification of the positive examples that the categorizer was able to classify. The higher its percentage, the fewer positive examples have been incorrectly classified.

### 5.5.5 F-Score or F-measure or F1
In an attempt to objectively deal with cases where a categorizer has disproportionately distributed classification errors, the metric F-Score was introduced, which is the harmonic mean between precision and recall and is calculated by the following relationship 38:

$$F_{Score} = \frac{2 \times recall \times precision}{recall + precision} = \frac{2TP}{2TP + FP + FN} \qquad (38)$$

The higher the percentage of the metric F-Score, the higher the respective two metrics.

In the following Figs 15, 16 and 17 are presented the classification reports plots by the SVM - Gaussian Kernel algorithm which produced the highest classification results.

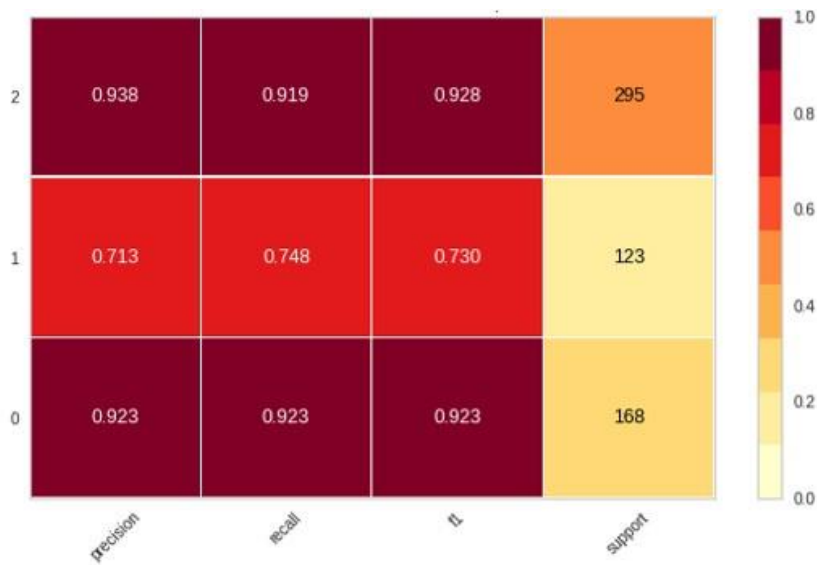

**Fig 15.** Classification report of the Row_Form_Bare dataset

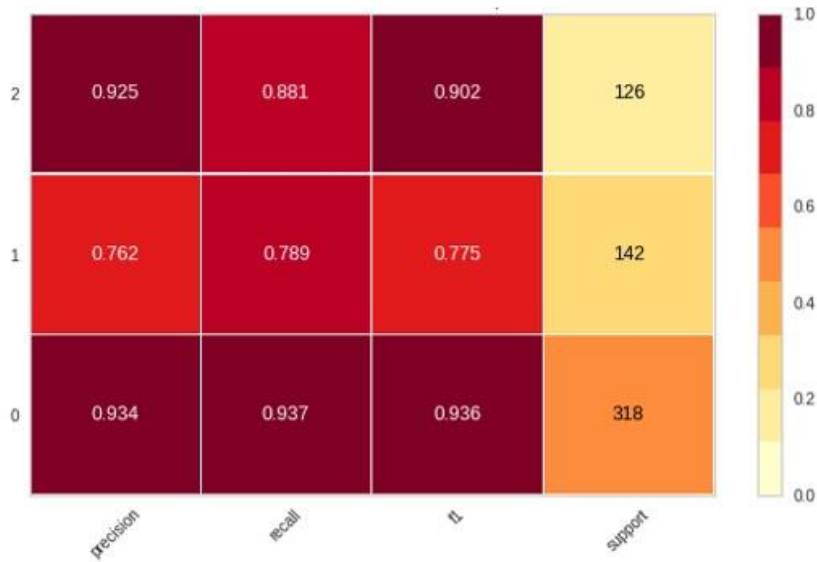

**Fig 16.** Classification report of the Row_Form_Full-Masonry dataset

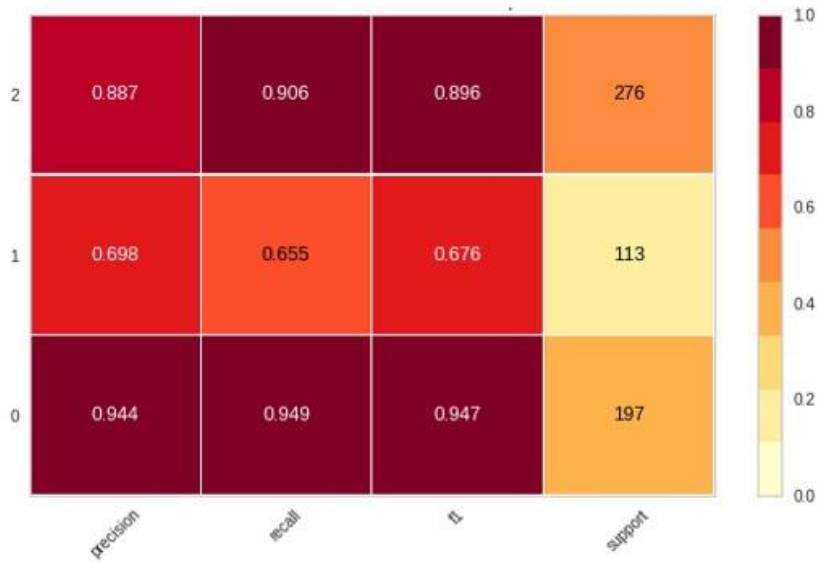

**Fig 17.** Classification report of the Row_Form_Pilotis dataset

Each classification report visualizer displays the precision, recall, F1, and support scores for the model. In order to support easier interpretation and problem detection, the report integrates numerical scores with a color-coded heatmap. All heatmaps are in the range (0.0, 1.0) to facilitate easy comparison of classification models across different classification reports.

### 5.5.6  Receiver Operating Characteristic (ROC)

This metric can be applied to categorizers that have as output trust. In this case, the categorizer predicts one class if its confidence for it exceeds a threshold. For the formation of the ROC Curve, various threshold values are used and the True Positive Rate (TPR) and False Positive Rate (FPR) percentages are noted for each of them. These value pairs are plotted on a graph where the *y*-axis corresponds to the TPR and the *x*-axis to the FPR. The performance of each categorizer is represented by a point on the ROC curve. The advantages of this metric are that it gathers information about the prediction quality of the categorizer for different threshold values and is also independent of the class imbalance in the data. The following Figs 18, 19 and 20 are presented the ROC curve plots by the SVM - Gaussian Kernel algorithm.

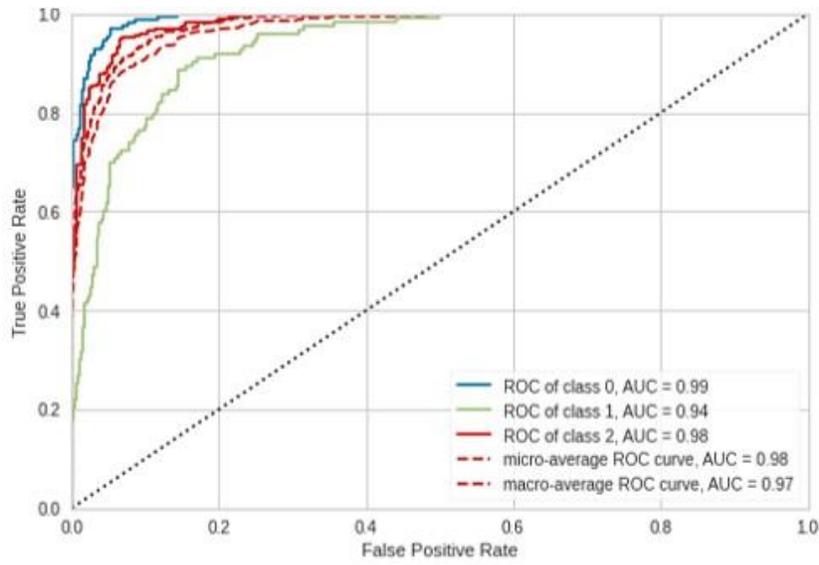

**Fig 18.** ROC curve of the Row_Form_Bare dataset

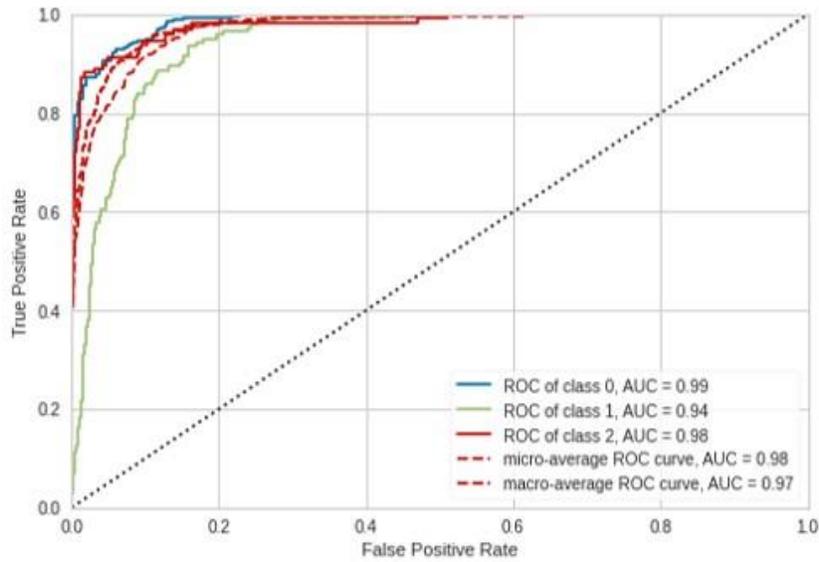

**Fig 19.** ROC curve of the Row_Form_Full-Masonry dataset

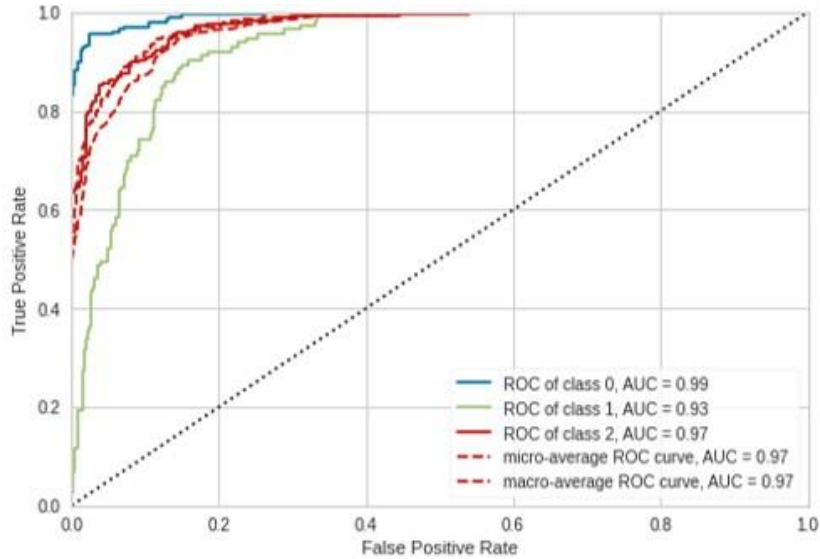

**Fig 20.** ROC curve of the Row_Form_Pilotis dataset

The above ROC curves are the measure of the SVM - Gaussian Kernel classifier predictive quality that compares and visualizes the tradeoff between the model's sensitivity and specificity. The higher the ROC, the better the model generally is. However, it is also important to inspect the "steepness" of the curve, as this describes the maximization of the true positive rate while minimizing the false positive rate.

### 5.5.7 Error Functions

Estimator function or estimator is a function of the random sample used to estimate an unknown parameter of a distribution function. The estimators in the case of classification refer to cost or error functions, which are able to quantify the classification variance achieved by an algorithm. The two cost or error functions used to categorize this method are the following:

1. Cohen's Kappa Statistics (CKS): This is a statistical measurement that provides information about the amount of agreement between the truth map and the final ranking map. It is the percentage agreement between two raters, where each classifies $N$ items into $C$ mutually exclusive categories. The definition of CKS calculate by the following equation 39:

$$\kappa = \frac{p_o - p_e}{1 - p_e} = 1 - \frac{1 - p_o}{1 - p_e} \qquad (39)$$

where $p_o$ is the relative observed agreement among raters (identical to accuracy), and $p_e$ is the hypothetical probability of chance agreement. The observed data are used to calculate the probabilities of each observer, to randomly see each category. More specifically, 0 = agreement equivalent to chance, 0.1 – 0.20 = slight agreement, 0.21 – 0.40 = fair agreement, 0.41 – 0.60 = moderate agreement, 0.61 – 0.80 = substantial agreement, 0.81 – 0.99 = near perfect agreement and 1 = perfect agreement.

2. Matthews Correlation Coefficient (MCC): MCC is used in machine learning as a measure of the quality of classifications. It is considered a balanced measure that can be used even if the sizes of the classes are very different, as it calculates a correlation coefficient value of [-1, +1] with +1 representing a perfect prediction, 0 an average random prediction and - 1 a reverse prediction. It is calculated by the formula 40:

$$\text{MCC} = \frac{\text{TP} \times \text{TN} - \text{FP} \times \text{FN}}{\sqrt{(\text{TP} + \text{FP})(\text{TP} + \text{FN})(\text{TN} + \text{FP})(\text{TN} + \text{FN})}} \qquad (40)$$

The class prediction error chart provides a way to quickly understand how good the classifier is at predicting the right classes. The following Class Prediction Error plots show the support (number of training samples) for each class in the fitted classification model as a stacked bar chart. Each bar is segmented to show the proportion of predictions (including false negatives and false positives, like the Confusion Matrix) for each class. We use each

Class Prediction Error plot to visualize which classes of the classifier is having a particularly difficult time with, and more importantly, what incorrect answers it is giving on a per-class basis. This enables us to better understand the strengths and weaknesses of each model and particular challenges unique to each dataset. The Figs 20, 21 and 22 are presented the Class Prediction Error plots by the SVM - Gaussian Kernel algorithm.

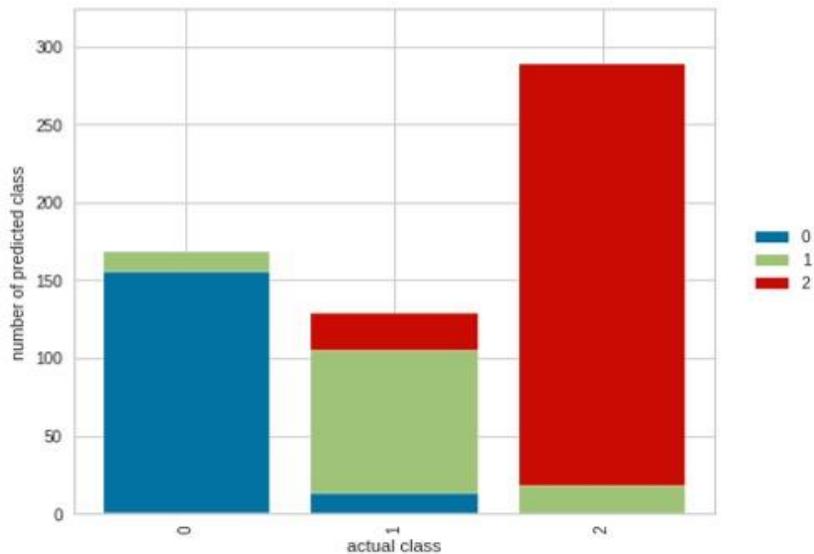

**Fig 21.** Class Prediction Error of the Row_Form_Bare dataset (for classes' definition see Table 1)

In the above example, while the classifier appears to be fairly good at correctly predicting classes 0 and 2 based on the features of the Row_Form_Bare dataset, it often incorrectly labels in class 1.

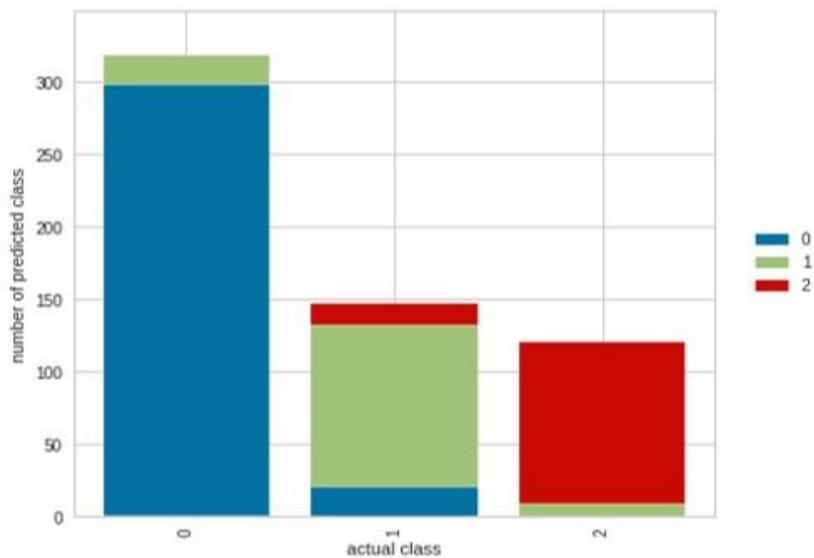

**Fig 22.** Class Prediction Error of the Row_Form_Full-Masonry dataset (for classes' definition see Table 1)

Similar, in the above example, the classifier it often incorrectly labels in class 1.

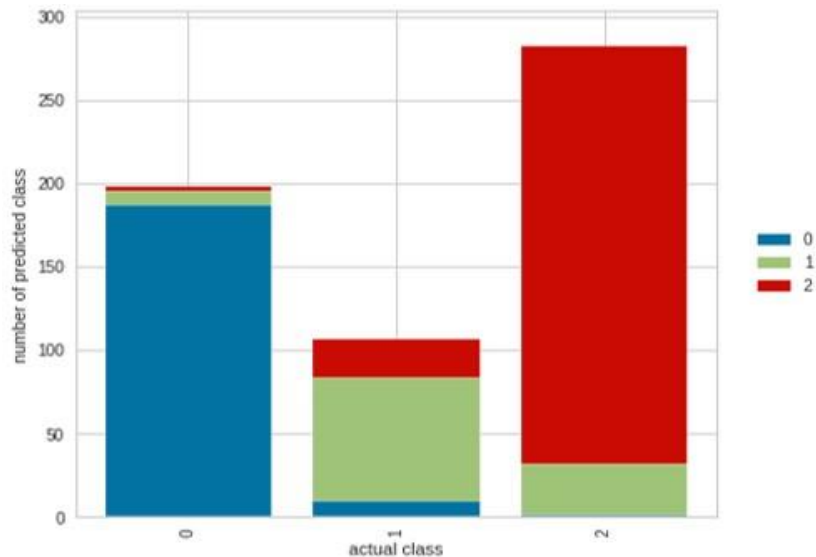

**Fig 23.** Class Prediction Error of the Row_Form_Pilotis dataset (for classes' definition see Table 1)

In the above example, the classifier appears to be fairly good at correctly predicting class 2. Also, it often incorrectly labels in class 0 and 1.

## 6. Conclusions

The present paper investigates the Machine Learning techniques' ability to reliably classify the seismic damage potential of R/C buildings. For this aim, a training dataset consisting of 30 3D R/C buildings with different structural parameters was chosen. The buildings were designed based on the provisions of EC8 and EC2. For each one of these buildings three different configurations as regards their masonry infills were considered (without masonry infills, with masonry infills in all stories and with masonry infills in all stories except for the ground story), leading to three different data subsets with 30 buildings each. Then, the buildings were analysed my means of the Nonlinear Time History Analyses method for 65 appropriately chosen real earthquake records. Both seismic and structural parameters widely used in the literature were selected as inputs in the process of Machine Learning methods. The quantification of the buildings' damage level was done by means of the well-documented Maximum Interstory Drift Ratio. As proved, the Machine Learning methods that are mathematically well-established and their operations that are clearly interpretable step by step can be used to solve some of the most sophisticated real-world problems in consideration.

The findings of the present work can be summarized as follows:

1. Different Machine Learning algorithms running on the same data set can lead to different predictions with little or no overlap between them and, also, different parameters of the same algorithm can affect the results of the buildings' damage assessment.
2. The SVM method used is not prone to overfitting to a specific data set compared to other methods and is also a robust method against data noise and outliers.
3. The important advantage of the SVM is that the optimization method that is used by the algorithm presents a total minimum, giving a unique optimal choice, which does not happen in other methods such as Neural Networks that can be trapped in local minima.
4. The performance of the SVM classifier depends not only on the algorithmic mechanism of the classifier itself (decision method) but also on the kernel to be applied to it, as its performance varies with the use of different kernels. Adjusting a kernel suitable to improve alignment with the samples of the training set which are labeled, significantly increases the fit with the samples of the test set, giving quite improved classification accuracy. Therefore, choosing the right kernel is a vital issue for the performance of the final classification model.
5. The SVMs have significant generalizability to non-linearly separable data by incorporating the kernel trick. By applying kernel functions it is possible to produce nonlinear models that lead to linearity in larger spaces. In addition, the number of parameters to be configured in SVMs is smaller compared to several corresponding methodologies. Also, for the classification of a new element in a class, the classification process is based only on the similarity of the element unknown to the algorithm and the most important elements of each class (support vectors), so the method reduces significantly the computational cost and the requirement resources.
6. The results confirm the need for further and in-depth exploration of the results related to the thorough evaluation of similarity measures and/or distance functions used in statistical approaches to implement features of the similar issues, in order to reduce the uncertainty in the decisions that include the data used to predict seismic damages.

Future work on this topic could focus on applying all of the above learning and prediction methods/models to different data sets, in order to assess whether they perform equally well on different data. This is also a way to test the insides of other methods of experts in terms of their robustness, i.e. to compare their performance for different data and to see if their accuracy is around the same percentages, so that they can be considered reliable as models. Also, ensemble methods could be applied with different combinations of individual classifiers, in terms of the number and nature of the latter, for further analysis of their function. In addition, other Machine Learning or Deep Learning models could be considered, such as the method based on the expectation-maximization algorithm, LSTM Neural Networks, etc. Finally, it will be important to look at different ways of selecting features, as well as mixing different selection methods, where it is expected very little overlap of insides indicators, in order to see if mixing methodology can statistically or mechanically lead to increased performance of the methods.